\documentclass[letterpaper, 10 pt, conference]{ieeeconf}  

\IEEEoverridecommandlockouts                              

\usepackage{graphics} 
\usepackage{epsfig} 
\usepackage{amsmath} 
\usepackage{float}
\usepackage{hyperref}

\usepackage{lipsum}
\usepackage{graphicx}

\usepackage[]{algorithm2e}
\usepackage{makecell}

\title{\LARGE \bf
Real-time Vision-based Depth Reconstruction with NVidia Jetson
}

\author{Andrey Bokovoy$^{1, 2}$, Kirill Muravyev$^{1, 3}$ and Konstantin Yakovlev$^{1, 3}$
\thanks{$^{1}$All authors are with the Artificial Intelligence Research Institute, Federal Research Center ``Computer Science and Control'' of Russian Academy of Sciences, Moscow, Russia.}%
\thanks{$^{2}$Andrey Bokovoy is also with the Peoples’ Friendship University of Russia (RUDN University), Moscow, Russia.}%
\thanks{$^{3}$Kirill Muravyev and Konstantin Yakovlev are also with the Moscow Institute of Physics and Technology, Dolgoprudny, Russia.}%
\thanks{Corresponding author is Andrey Bokovoy:
{\tt\small bokovoy@isa.ru}%
\newline
\newline camera-ready version as submitted to ECMR 2019}%
}

\begin{document}

\maketitle
\thispagestyle{empty}
\pagestyle{empty}

\begin{abstract}

Vision-based depth reconstruction is a challenging problem extensively studied in computer vision but still lacking universal solution. Reconstructing depth from single image is particularly valuable to mobile robotics as it can be embedded to the modern vision-based simultaneous localization and mapping (vSLAM) methods providing them with the metric information needed to construct accurate maps in real scale. Typically, depth reconstruction is done nowadays via fully-convolutional neural networks (FCNNs). In this work we experiment with several FCNN architectures and introduce a few enhancements aimed at increasing both the effectiveness and the efficiency of the inference. We experimentally determine the solution that provides the best performance/accuracy tradeoff and is able to run on NVidia Jetson with the framerates exceeding 16FPS for $320 \times 240$ input. We also evaluate the suggested models by conducting monocular vSLAM of unknown indoor environment on NVidia Jetson TX2 in real-time. Open-source implementation of the models and the inference node for Robot Operating System (ROS) are available at \url{https://github.com/CnnDepth/tx2_fcnn_node}. 

\end{abstract}

\section{INTRODUCTION}

Depth reconstruction (estimation) is one of the important problems in mobile robotics, augmented reality, computer aided design etc. The sensors that explicitly provide range measurements such as LIDARs, RGB-D cameras etc., are typically i) expensive, ii) large and heavy, iii) power-demanding, which prevents their widespread usage especially when it comes down to compact mobile robots (like small drones). Thus a strong interest exists in depth estimation using a single camera, as almost every mobile robot is equipped with this sensor. Moreover, there exist data-driven learning-based approaches that are capable of solving monocular vision-based depth reconstruction tasks with suitable (for typical mobile robotics applications) accuracy -- see works \cite{garg2016unsupervised, li2015depth, godard2017unsupervised}. Commonly, the main focus of such papers is increasing the accuracy while the performance issues are left out of scope. As a result, the majority of the state-of-the-art methods for depth reconstruction are very resource demanding and need high-performance graphic processing units (GPU) to work in real time. Thus, they are not suitable for creating a fully-autonomous robotic system equipped with a typical embedded computer, even the one that is particularly suitable for image processing with neural networks -- NVidia Jetson TX2. On the other hand, there are plenty of reports of this embedded computer being successfully used for autonomous navigation, SLAM etc., but still there is a limited number of papers, e.g. \cite{spek2018cream}, that report a successful usage of single camera deep-learning driven depth estimation that works in real-time on NVidia Jetson TX2. Furthermore, to the best of our knowledge, there are no reproducible results (in terms of open-source code) of FCNNs for real-time embedded vSLAM usage. The foregoing defines the scope of this work. We wish to present a CNN-based depth reconstruction method that i) is accurate enough to be used within the monocular vSLAM pipeline and is equivalent accuracy-wise to the state-of-the-art, ii) is fast enough to work in real time on NVidia Jetson TX2, iii) is open to the community, i.e. comes with a source-code of the ROS-node.

\begin{figure}[t]
    \centering
    \includegraphics[width=0.48\textwidth]{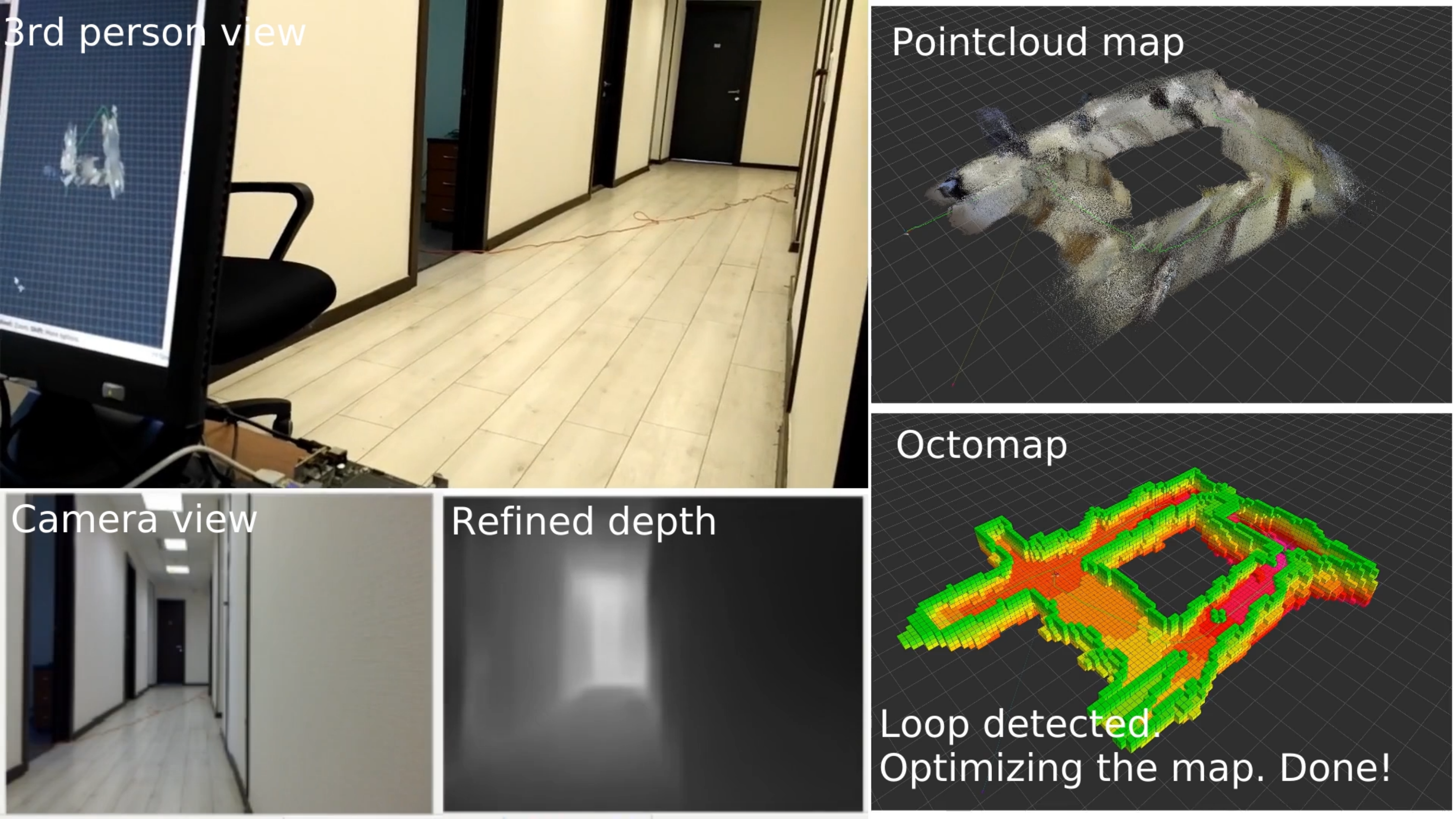}
    \caption{Monocular vSLAM based on FCNN depth reconstruction and running in real time on NVidia Jetson TX2. This is a screenshot of the video available at: \url{https://youtu.be/ayjvfzm-C7s} }
    \label{figurelabel1}
\end{figure}

\section{Related work}
\label{sec2_problem_statement}

\subsection{Depth reconstruction from single image}
\label{subsec2.1}

Depth estimation from single image has a prolonged history of studies. Initially such techniques as image pre-processing, feature extraction, edge detection etc. were widely utilized to solve the task. In \cite{levin2007image} authors use hardware modification of the camera's lens to make simultaneous image and depth extraction with cost-efficient algorithm. Proposed method exploits the prior knowledge about real images, particularly their statistical distribution \cite{olshausen1996natural}. However, in most cases, resultant depth maps require manual correction.

In \cite{saxena20083} Markov Random Field (MRF) is used for patch-based depth reconstruction with single image. The original image is divided into a set of patches in different scales. The hand-crafted features are then applied to these patches. Using the statistics from different scales of the patches, MRF models the relationship between the depth of the patch and the neighborhood patches, reconstructing the depth of the whole image. 

Not only the knowledge about image statistics may provide the ability to reconstruct the depth map of the single image, but the prior information about the environment. For example, in \cite{delage2006dynamic}
the information that indoor environment mostly consists of vertical and horizontal lines (walls, floor and etc.) is utilized. This helps to determine the perspective and reconstruct the depth in scenarios where, for example, the robot moves along the corridor. However, it fails in other types of environments.

Recently, deep learning and convolutional neural networks became tools of choice for depth reconstruction as they significantly outperform methods based on hand-crafted features. Fully-convolutional neural networks \cite{long2015fully}, consisting of encoder and decoder, are the most common architectures to be used for depth reconstruction.

One of the pioneer works in CNN-based depth estimation from single image is \cite{eigen2014depth}. Authors use coarse-scale network to predict the depth of the image at global level (as a dense depth map). Then, local fine-scale network aligns the map with image's local details, such as objects or wall edges.         

\begin{figure*}[ht]
  \includegraphics[width=\textwidth]{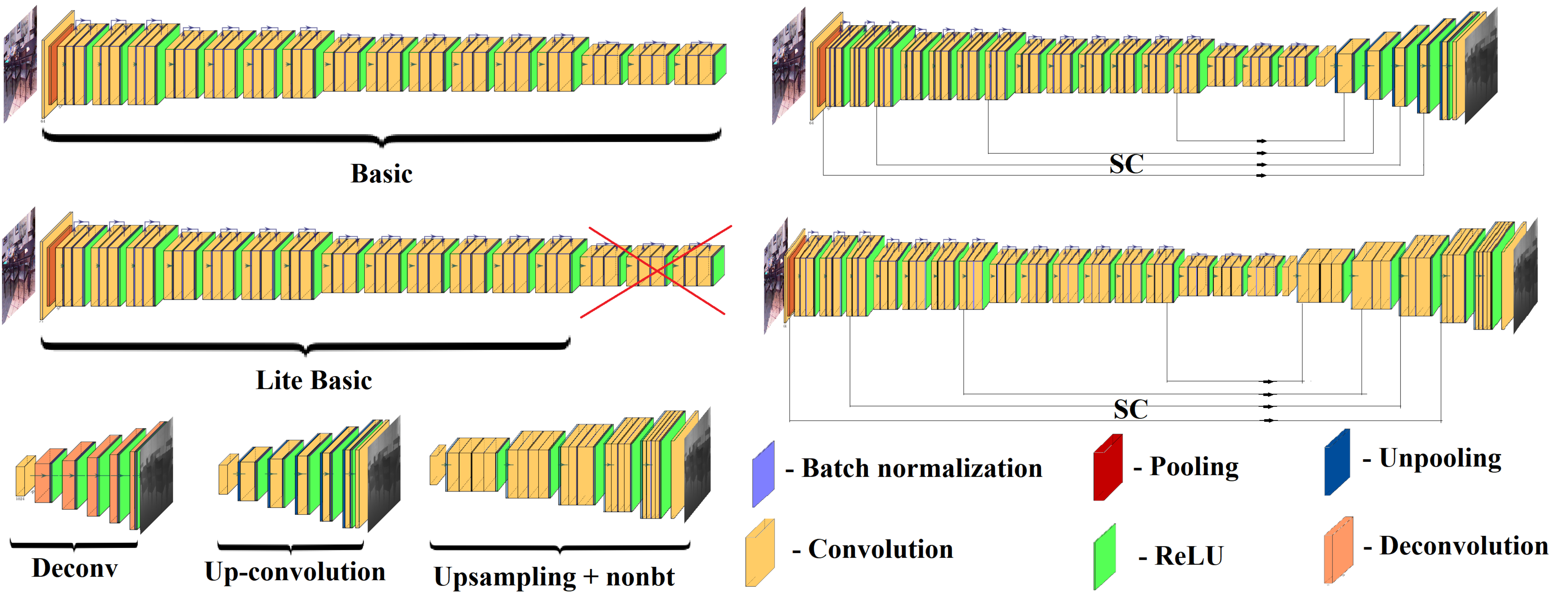}
  \caption{Visualization of evaluated network architectures.}
  \label{figurelabel4}
\end{figure*}

In \cite{garg2016unsupervised} authors utilize inverse depth maps, produced by neural network, with known inter-view displacement to achieve unsupervised learning of the CNN, solving the problem of collecting large datasets for training. The training set consists of only 22 600 stereo images, without any data augmentation or usage of pretrained decoder (in this particular case - AlexNet \cite{krizhevsky2012imagenet}). 

In \cite{laina2016deeper} the original up-convolution algorithm is proposed as well as the reverse Huber loss function \cite{owen2007robust}. This architecture was tested for real-time applications, but the targeted GPU is NVidia GeForce GTX TITAN with 12GB of GPU memory, which is more powerful compared to NVidia Jetson. Notably, this FCNN is used within the SLAM pipeline presented in \cite{tateno2017cnn} on desktop PC with Intel Xeon CPU at 2.4GHz with 16GB of RAM and a Nvidia Quadro K5200 GPU with 8GB of VRAM. 

Authors of MegaDepth \cite{li2018megadepth} and DenseDepth \cite{alhashim2018high} focus mainly on improvement of FCNN learning phase. In \cite{li2018megadepth} an original loss function is proposed as well as learning strategy and data augmentation techniques. MegaDepth authors focus on improving the quality of the training dataset. They also suggest routines for depth refinement with automatic ordinal labeling and semantic segmentation. Both architectures are too heavy for real-time image processing, especially on embedded systems.

In general, vast majority of the FCNN-depth-reconstruction papers leaves computing constraints out of the scope. We, in contrary,  wish to focus on real-time performance on embedded computers that are widely used in modern robotics.

\subsection{Depth reconstruction on embedded systems}

One of the most popular embedded computer nowadays for robotics is Raspberry Pi. It is used for autonomous car navigation \cite{pannu2015design}, grid-based path planning \cite{andreychuk2018empirical}, object detection \cite{pereira2014low} etc. Despite the low-power CPU and the absence of GPU, Raspberry is even suitable for some CNN-based tasks. E.g. in \cite{pena2017benchmarking} object recognition was considered and simple and light architectures were used on par with the Movidius Neural Compute Stick hardware acceleration. Framerates of 0.5-1.5 FPS were achieved. In this work we are targeting framerates of $>$5FPS.

The next popular and more powerful embedded system for robotics is ODROID. In \cite{forster2014svo}\cite{faessler2016autonomous} it was reported to be used as an on-board computer for semi-dense visual odometry on small quadrotor. Framerate of 5FPS was achieved. In general, the lack of acceleration tools for ODROID forces usage of external hardware for CNN processing. 

Finally, some of the platforms were produced recently with deep learning tasks in mind. For example, NVidia Jetson is used for real-time GPU accelerated image processing tasks, including FCNN semantic segmentation \cite{paszke2016enet}, object detection \cite{shafiee2017fast}, image classification \cite{likamwa2016redeye} etc. Low power consumption and compact size of Jetson make this computer perfectly suitable for mobile robotics and the availability of GPU makes it the research platform of our choice. There is a successful report of running FCNN for depth reconstruction using NVidia Jetson TX2 in real-time \cite{spek2018cream}. In this work the authors introduce a light-weight encoder-decoder architecture, that was trained with knowledge transfer from a more heavy one. They achieved 30 FPS inference with comparable to state-of-the-art accuracy. However, the vSLAM application is not well-studied (only the scale-drift problem) and the results are not reproducible (in terms of code, TensorRT engine's binaries and etc.). In this work we provide an open-source solution for depth reconstruction and vSLAM.

\section{Evaluated architectures}
\label{sec3_}

With the focus on on real-time performance under limited computational resources we study different variations of Fully-convolutional Neural Networks for depth reconstruction (see Fig. \ref{figurelabel4}).

Stereotypical FCNN model for depth reconstruction consists of the encoder and the decoder. The former extracts the high-level features from the input image while the latter generates the depth maps from these features. We used ResNet50 (and its cropped version) as the encoder and a few different decoders. On top of that we enhance some blocks of the network to make it work faster while keeping the accuracy at the appropriate level. By combining different encoders, decoders and enhancements we end up with 6 different architectures to evaluate.

\subsection{Encoder}
ResNet50 \cite{he2016deep} is known to be versatile feature extractor, so we chose it as the encoder. Despite being a deep (50 layers) network with several residual blocks, it's fast enough to operate in real-time. The output of standard ResNet50 for 640x480x3 input is 20x15x2048 feature maps. We further denote this encoder as \textbf{Basic}. 

We also evaluate a light version of ResNet50 which lacks the last stack of residual blocks, so the output for 640x480x3 image is 30x40x1024. This greatly increases the performance of the network while keeping the accuracy at the appropriate level (i.e. sufficient for vSLAM purposes). The cropped version of ResNet50 is referred as \textbf{Lite Basic}.

\subsection{Decoder}

As the baseline decoder we use the one that is composed of 5 deconvolution blocks (\textbf{Deconv}). Each block consists of 5x5 deconvolution + batch normalization + activation (ReLU).

Second, we evaluate the decoder that is composed of the blocks that use the upsampling followed by the non-bottleneck convolution followed by the ReLU. After 5 such blocks we reduce the output to the one channel, i.e. -- depth, with $5 \times 5$ convolution. The upsampling is performed with the nearest neighbour algorithm. It is followed by the two $3 \times 3$ convolutions implemented as the factorized non-bottleneck block, suggested in \cite{romera2018erfnet}. It substitutes the conventional $3 \times 3$ convolution with a series of $3 \times 1$ and $1 \times 3$ convolutions that results in faster inference. We denote this decoder as \textbf{Upsampling + nonbt}.

Third, we use the \textbf{up-convolution} decoder. Each block of this decoder is composed of unpooling + $5 \times 5$ convolution + batch normalization + ReLU + Dropout. In total 5 blocks are stacked when the \textbf{Basic} encoder is used, 4 -- in case of \textbf{Lite Basic}.

\begin{figure}[t]
    \centering
    \includegraphics[width=0.35\textwidth]{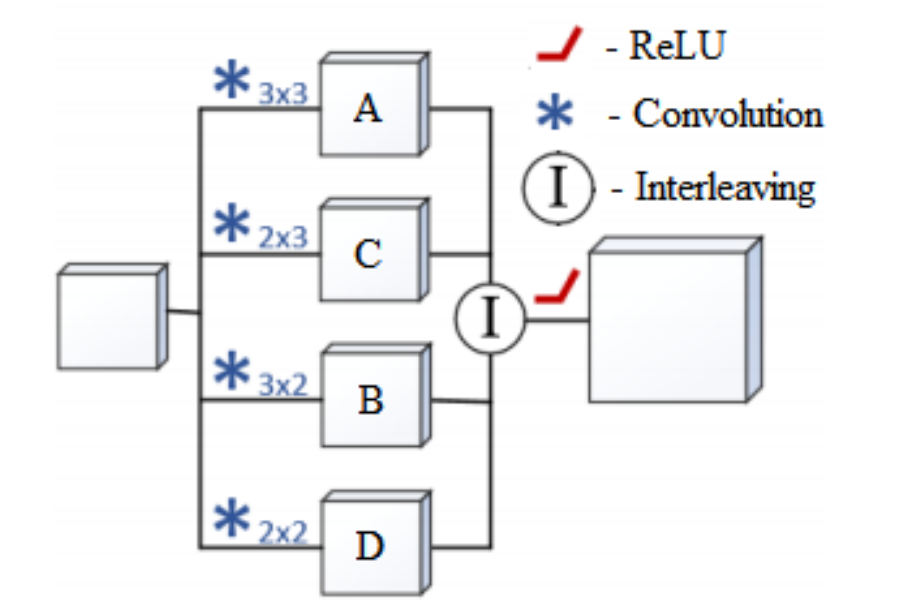}
    \caption{Faster up-convolution block architecture.}
    \label{figurelabel5}
\end{figure}

It was shown in \cite{laina2016deeper} that one can substitute the unpooling + $5 \times 5$ convolution with the smaller sized convolutions that are interleaved into the resultant feature map -- see Fig. \ref{figurelabel5}. Such approach leads to a faster inference although splitting the $5\times 5$ convolution into 4 parts leads to inconsistent gradients and worse weights optimization during training. To overcome this we suggest using original up-convolution decoder during learning and then transferring the learned weights to the faster architecture (the one that utilizes interleaving). We denote \textbf{interl} the unpooling decoder that relies on interleaving at both learning and inference. We denote \textbf{interl + T} the decoder with the original (non-interleaved) unpooling + convolution layers used for training and interleaved layers (with the transferred weights) for inference.

\subsection{Shortcuts}

We also use shortcuts (or skip connections, referred as \textbf{SC}) as the projection from encoder layers to decoder. Despite it does not always improves model's accuracy, it produces more sharpened depths on the objects' edges. For different combinations of encoder-decoder architectures we use different shortcuts. For \textbf{Basic + SC} encoder with \textbf{Upsampling + nonbt} decoder, we connect the output of last convolution block in every stack of the encoder with respective outputs of decoder blocks (see Fig.\ref{figurelabel4}). For \textbf{Basic + SC + interl} there are 2 shortcuts that connects the outer and middle blocks of encoder and decoder. 

\subsection{Interleaving implementation for faster up-convolution}
Interleaving is an approach proposed in \cite{laina2016deeper} that substitutes un-poolling + $5 \times 5$ convolution with 4 convolutions which outputs are interleaved into a single feature map. This operation is equivalent in terms of the resultant output but is more computationally efficient. 

Our implementation of interleaving differs from the one residing in the authors' repository\footnote{\url{https://github.com/iro-cp/FCRN-DepthPrediction}} in the following way: while authors of the original method apply interleaving as a 3-step operation, merging first and second pairs of weights, then merging the results together, we do it in a single operation (see Alg. 1) with respect to multithreading, saving CPU to GPU memory copy time\footnote{Code for Tensorflow and TensorRT is available at \url{https://github.com/CnnDepth/interleave_op}.}.

\begin{algorithm}
 \KwData{A, B, C, D - input 4D tensors in (N,H,W,C) format, N - batch size, H - image height, W - image width, C - number of channels}
 \KwResult{Out - interleaved output 4D tensor in (N,H,W,C) format }
 \For{$i\gets indexOfElementInBlock $}{
  $n_{in} = i / (H * W * C)$; \\
  $h_{in} = ( i \mod (H * W * C) ) / ( W * C )$; \\
  $w_{in} = (i \mod (W*C) / C))$; \\
  $c_{in} = (i \mod C)$; \\
  $index_{in} = n_{in} * H * W * C / 4 + (h_{in}/2)* W * C / 2 + (w_{in}/2)* C + c_{in}$;\\
  \eIf{$h_{in} is even$}{
    \eIf{$w_{in} is even$}{
      $Out[i] = A[index_{in}]$;\\
    }{
      $Out[i] = B[index_{in}]$;\\
    }
  }{
    \eIf{$w_{in} is even$}{
      $Out[i] = C[index_{in}]$;\\
    }{
      $Out[i] = D[index_{in}]$;\\
    }
  }
 }
\caption{Interleaving implementation}
\end{algorithm}

\subsection{Loss functions}

Running preliminary experiments we discovered, that errors on near and far predicted depths deviate significantly from the mean error. The far-away pixels are less important in the context of autonomous navigation of a mobile robot \cite{1389826}, but the incorrect estimation of depths in the vicinity of the camera may lead to undesirable outcomes (e.g. crashing into the obstacle). To mitigate this issue we suggest using 2 original loss functions.

The first one is a combination of two losses: 

\begin{center}
$MSE+REL = \alpha_1 \cdot MSE + \alpha_2 \cdot REL$,


$MSE=\dfrac{1}{H \cdot W} \sum\limits_{i=1}^H \sum\limits_{j=1}^W (D_{ij}^{\mbox{*}} - D_{ij})^2$,

$REL=\dfrac{1}{H \cdot W} \sum\limits_{i=1}^H \sum\limits_{j=1}^W \bigg( 1 - \dfrac{D_{ij}}{D_{ij}^{\mbox{*}}} \bigg)^2$,

\end{center}

$D_{ij}^{\mbox{*}}$ -- ground-truth pixel depth, $D_{ij}$ -- predicted pixel depth, $\alpha_1, \alpha_2$ -- user-defined parameters (we set them to 1 and 2 respectively in our experiments).

The second component of that loss (REL) accounts for the fact that, e.g. the 0.5m error at the distance of 1m is worse than the 0.5m error at the distance 10m. 

Another loss function we used is the the BerHu loss \cite{li2015depth}:

\begin{center}
$L(D_{ij}^{\mbox{*}}, D_{ij}) = \begin{cases}
	|D_{ij}^{\mbox{*}} - D_{ij}|, \text{\ \ \ \ \ \ \ \ \ \ } |D_{ij}^{\mbox{*}} - D_{ij}| < k\\
	\dfrac{(D_{ij}^{\mbox{*}} - D_{ij})^2 + k^2}{2 k}, \text{\ } |D_{ij}^{\mbox{*}} - D_{ij}| \geq k\\
	\end{cases} $
	
$BerHu = \dfrac{1}{H \cdot W} \sum\limits_{i=1}^H \sum\limits_{j=1}^W L(D_{ij}^{\mbox{*}}, D_{ij})$.
\end{center}

BerHu loss accounts for the same proposition -- the model should be more sensitive to the errors within the close range. It needs to be provided with the threshold, $k$, that accounts for what is ``near'' and what is ``far'' and that threshold is fixed during learning. On contrast, we suggest to alter the value of $k$ in the following fashion. At each step during learning phase, we additionally compute two BerHu losses over the pixels lying at depths $[k - \delta; k]$ and $[k; k + \delta]$ and compare them numerically. Then $k$ is shifted towards bigger mean error by $k \pm lr * \delta$.\footnote{Code for adaptive BerHu loss is available at \url{https://github.com/CnnDepth/fcrn_notebooks}.}. This allows us to adaptively adjust the closeness threshold while training. In our work, variables $\delta$ and $lr$ were set to 1 and 0.01 respectively as initial values. We refer to this function as \textbf{aBerHu}.

\tabcolsep=0.11cm
\begin{table*}[ht]
\caption{Evaluation of loss function for network architectures presented in Section \ref{sec3_}. The values are those originally reported by the authors in their respective paper.}
\label{table_evaluation}
\begin{center}
\begin{tabular}{|c||c c c||c c|c c c||c|c|}
\hline

\hline Name & Loss & Input resolution & Decoder & MSE & REL & $\delta^1$ & $\delta^2$ & $\delta^3$ & \makecell{PC \\ time (s)} & \makecell{Jetson\\ time (s)}\\
\hline
Wang et al. \cite{wang2015towards} & Custom & - & - & 0.555 & 0.220 & 0.605 & 0.890 & 0.970 & - & -\\
Eigen et al. \cite{eigen2014depth} & Custom & 304x228 & - & 0.823 & 0.215 & 0.611 & 0.887 & 0.971 & - & - \\
Laina et al. \cite{laina2016deeper} & BerHu & 304x228 &  - & 0.328 & 0.127 & 0.811 & 0.953 & 0.988 & - & -\\
Alhashim et al. \cite{alhashim2018high} & Custom & 640x480 &  - & 0.152 & 0.103 & 0.895 & 0.980 & 0.996 & - & -\\

\hline
Basic & BerHu & 640x480 &  Deconv & 0.467 & 0.186 & 0.718 & 0.929 & 0.980 & 0.144 & 0.152\\
Basic + SC & BerHu & 640x480 & Deconv & 0.487 & 0.194 & 0.695 & 0.915 & 0.975 & 0.521 & 0.563\\
Basic + SC & \makecell{aBerHu} & 640x480 & Upsampling + nonbt & 0.440 & 0.184 & 0.725 & 0.932 & 0.982 & 0.158 & 0.215\\
Basic + SC & MSE + REL & 640x480 & Upsampling + nonbt & 0.419 & \textbf{0.173} & \textbf{0.748} & \textbf{0.944} & \textbf{0.987} & 0.158 & 0.215 \\
Basic + SC & MSE + REL & 320x240 & Upsampling + nonbt & \textbf{0.408} & 0.180 & 0.746 & 0.940 & 0.984 & 0.049 & 0.062 \\
Lite Basic + SC & MSE + REL & 320x240 & Upsampling + nonbt & 0.533 & 0.202 & 0.687 & 0.915 & 0.979 & 0.035 & 0.049 \\
Basic + SC + interl & MSE + REL & 640x480 & Up-convolution & 0.514 & 0.206 & 0.708 & 0.912 & 0.970 & 0.285 & 0.328\\
Basic + SC + interl & MSE + REL & 320x240 & Up-convolution & 0.580 & 0.215 & 0.673 & 0.899 & 0.965 & 0.057 & 0.067\\
Basic + SC + interl + T & MSE + REL & 640x480 & Up-convolution & 0.445 & 0.178 & 0.714 & 0.939 & \textbf{0.987} & 0.181 & 0.227\\
Basic + SC + interl + T & MSE + REL & 320x240 & Up-convolution & 0.495 & 0.181 & 0.724 & 0.940 & 0.983 & 0.048 & 0.061\\
Lite Basic + interl + T & MSE + REL & 640x480 & Up-convolution & 0.658 & 0.233 & 0.642 & 0.886 & 0.964 & 0.101 & 0.135\\
Lite Basic + interl + T & MSE + REL & 320x240 & Up-convolution & 0.660 & 0.236 & 0.649 & 0.881 & 0.960 & \textbf{0.027} & \textbf{0.037}\\
\hline
\end{tabular}
\end{center}
\end{table*}

\section{Experimental results}
\label{sec4_}

The considered FCNN architectures were implemented using Tensorflow \cite{abadi2016tensorflow} + Keras \cite{gulli2017deep} frameworks in Python for. Custom interleaving and up-convolution layers were implemented both for CPU and GPU using C/C++ with g++ and nvcc compilers respectively. Learning was performed on the Hybrid high-performance computing cluster of Federal Research Center Computer Science and Control of Russian Academy of Sciences. 

For inference tests we consider 2 possible scenarios: 1) fully autonomous depth reconstruction on NVidia Jetson TX2, 2) remote depth reconstruction with mobile PC (laptop). For both scenarios we use accelerated TensorRT framework. All the inference-related part was written in C/C++ using tools available in JetPack software package. All models and respective learned weights are converted to inference engine used by TensorRT. Our PC platform specification is as follows: Intel Core i7 8550, 20GB of RAM, NVidia MX150 GPU with 4GB of memory. 

\subsection{Dataset}

NYU dataset v2 \cite{Silberman:ECCV12} was used. It consists of more that 400 000 image-depth pairs taken from more then 470 scenes. All raw images were aligned with corresponding depth maps and pre-processed with bilateral filter to fill the missing depth values on the edges of objects. Since there are a lot of similar images in NYU Dataset, we performed random crop for each image-depth pairs, as well as random mirroring and rotations, which led to diversification of dataset. 

\subsection{Error metrics}

We used the following metrics to measure the accuracy of the depth estimation:

\begin{itemize}
    \item MSE -- mean squared error;
    \item REL -- mean relative error;
    \item Threshold accuracy $\delta^i$ -- \% of predicted depths $D_i\mbox{*}: \dfrac{1}{N} \sum_{i=1}^{N} max\bigg(\dfrac{D_i\mbox{*}}{D_i},\dfrac{D_i}{D_i\mbox{*}}\bigg)<\delta^i, \delta = 1.25$ 
\end{itemize}

\subsection{Results}

\begin{figure}[t]
    \centering
    \includegraphics[width=0.48\textwidth]{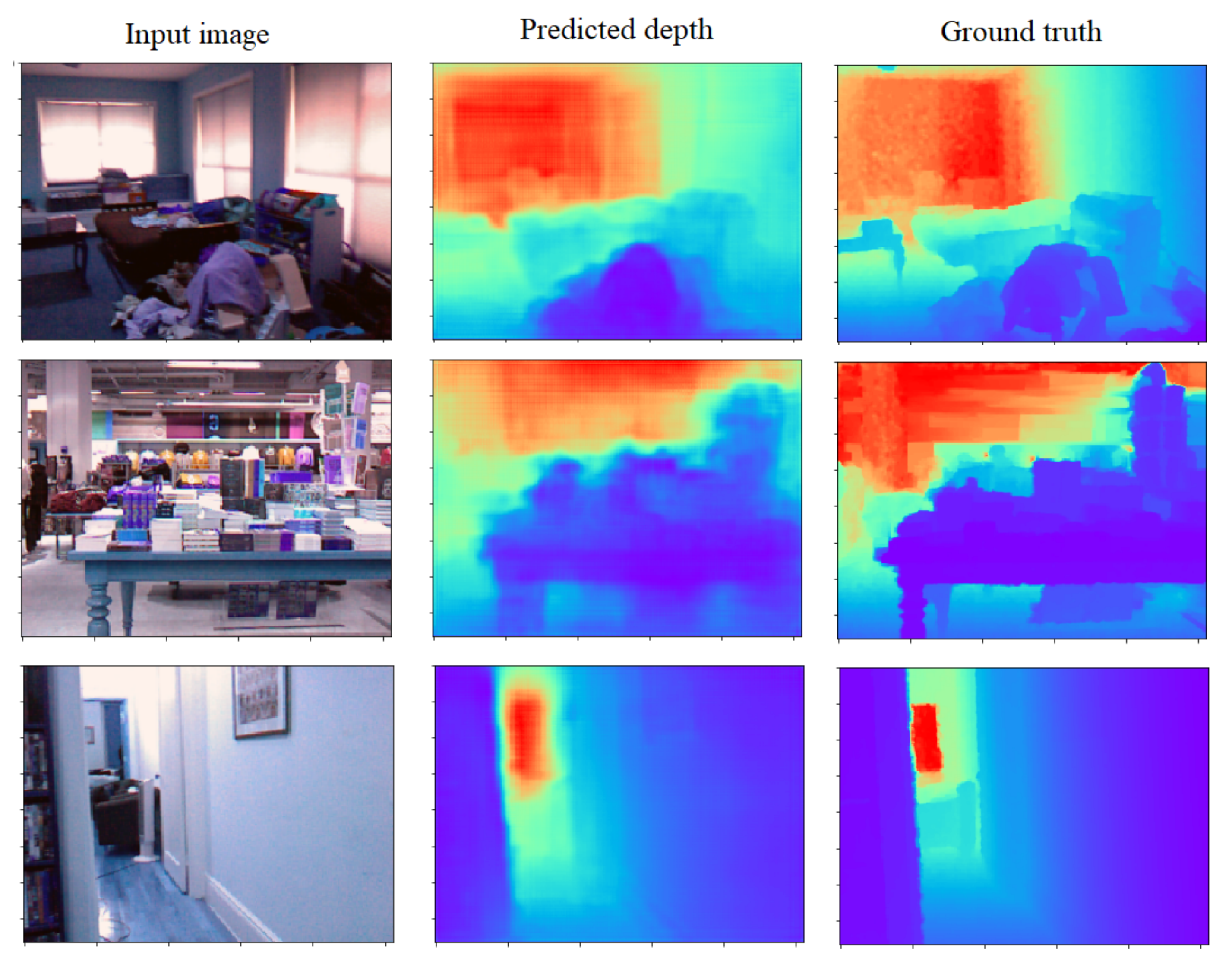}
    \caption{Visualization of introduced FCNN on NYU Dataset v2.}
    \label{figurelabel11}
\end{figure}

The results are presented in Table. \ref{table_evaluation}. As the best result, we achived 37ms per image inference on NVidia Jetson TX2 for \textbf{Lite Basic + interl + T} architecture. The inference speed is comparable to the state-of-the art method for depth-reconstruction with embedded platforms from \cite{spek2018cream}, while the accuracy is slightly worse, but applicable for real-time vSLAM purposes. On the other hand, we've achieved better REL error metrics then some basic architectures presented in \cite{laina2016deeper}, \cite{liu2016learning} and \cite{eigen2014depth}, that are focused on offline depth reconstruction, with \textbf{Basic + SC} encoder and \textbf{Upsampling + nonbt} decoder. Our tests showed, that even 62ms is enough for real-time vSLAM, but odometry may fail during fast translations due to low update rate of the camera images.

Evidently, using the \textbf{Lite Basic} encoder gives a notable inference speed improvement, but at the cost of the reduced accuracy.

\subsection{vSLAM evaluation}

We implemented the node for Robot Operating System (ROS) for FCNN inference. For localization and mapping we use RTAB-Map \cite{labbe2019rtab}. We run both FCNN inference and RTAB-Map on NVidia Jetson TX2. Our experiments show that developed networks are well suited for accurate and fast single-camera simultaneous localization and mapping on embedded platform. The model that provides the best performance/accuracy trade-off is \textbf{Basic + SC} encoder paired with \textbf{upsampling + nonbt} decoder. It runs on NVidia Jetson's GPU with 16 FPS, while vSLAM operates on CPU, so both algorithms don't interfere (the overall performance is approx. 12FPS). We open-source our FCNN inference implementation with all engines compiled for NVidia Jetson: \url{https://github.com/CnnDepth/tx2\_fcnn\_node}. The pipeline may be extended with other methods and algorithms available for ROS.

As shown in Fig. \ref{figurelabel1}, our FCNN inference + RTAB-Map is able to produce high-detailed maps of unknown environment with only single camera. In our previous work \cite{bokovoy2018sparse} we used state-of-the-art feature-based SLAM approach with the enhanced post-processing and still were not able to obtain maps of such quality. The video of the vSLAM evaluation is available at \url{https://youtu.be/ayjvfzm-C7s}.

\section{CONCLUSION AND FUTURE WORK}
\label{sec5_conclusion}
In this work we have evaluated different FCNN architectures for depth-reconstruction both on PC and NVidia Jetson TX2. We demonstrated, that the proposed models are able to run in real-time with the accuracy comparable to the state-of-the-art. We implemented the proposed methods as a part of ROS framework and tested it with RTAB-Map for the indoor SLAM. The results show that our pipeline is able to produce dense maps on embedded computer in real time. 

In future we wish to more thoroughly evaluate the proposed FCNNs within the vSLAM pipeline (e.g. map and pose accuracy estimation), and further use the produced maps for autonomous navigation, e.g. for path planning.

\section*{ACKNOWLEDGMENT}

This work is supported by the RSF project \#16-11-00048 (developing FCNN architectures and evaluating them) and by the ``RUDN University Program 5-100'' (post-processing of the experimental results).

\addtolength{\textheight}{-12cm}   

\bibliographystyle{IEEEtran}
\bibliography{IEEEexample.bib}

\end{document}